\documentclass[runningheads]{llncs}
\usepackage{graphicx}

\usepackage{amsmath}
\usepackage{authblk}
\usepackage{booktabs}
\usepackage{algorithm}
\usepackage{algorithmic}
\usepackage{amssymb}
\usepackage{subfig}
\usepackage{multirow}
\usepackage{footnote}
\usepackage{color}
\usepackage{tabularx}
\usepackage{nameref}
\begin{document}
\title{FlexNER: A Flexible LSTM-CNN Stack Framework for Named Entity Recognition}
\titlerunning{FlexNER}
%
\author{Hongyin Zhu\inst{1,2} \and
Wenpeng Hu\inst{3} \and
Yi Zeng\inst{1,2,4,5,}\thanks{Corresponding author}}

\authorrunning{Zhu et al.}

\institute{Institute of Automation, Chinese Academy of Sciences, Beijing, China \and
University of Chinese Academy of Sciences, Beijing, China \and
School of Mathematical Sciences, Peking University, Beijing, China \and
Center for Excellence in Brain Science and Intelligence Technology,
Chinese Academy of Sciences, Shanghai, China \and
National Laboratory of Pattern Recognition, Institute of Automation, Chinese Academy of Science, Beijing, China \\ 
\email{\{zhuhongyin2014,yi.zeng\}@ia.ac.cn, wenpeng.hu@pku.edu.cn}
}

%
\maketitle              
\begin{abstract}
Named entity recognition (NER) is a foundational technology for information extraction. This paper presents a flexible NER framework\footnote{\url{https://liftkkkk.github.io/FLEXNER/}} compatible with different languages and domains. Inspired by the idea of distant supervision (DS), this paper enhances the representation by increasing the entity-context diversity without relying on external resources. We choose different layer stacks and sub-network combinations to construct the bilateral networks. This strategy can generally improve model performance on different datasets. We conduct experiments on five languages, such as English, German, Spanish, Dutch and Chinese, and biomedical fields, such as identifying the chemicals and gene/protein terms from scientific works. Experimental results demonstrate the good performance of this framework.

\keywords{Named entity recognition  \and data augmentation \and LSTM-CNN.}
\end{abstract}
\section{Introduction}
The NER task aims to automatically identify the atomic entity mentions in textual inputs. This technology is widely used in many natural language processing pipelines, such as entity linking, relation extraction, question answering, etc. 
This paper describes a portable framework that can use different layer stacks and sub-network combinations to form different models. This framework does not rely on language/domain-specific external resources so it can reduce coupling.

While state-of-the-art deep learning models \cite{ma2016end,lample2016neural,collobert2011natural} resolve this problem in the sequence labeling manner, their models are usually trained on a fixed training set where the combination of entity and context is invariant, so the relationship between entity and context information is not fully exploited. Intuitively, adding diverse training samples \cite{jiang2014self,zhou2017neural} is helpful to train a better model, but expanding the existing training data is expensive. The Wikipedia entity type mappings \cite{ni2017improving} or the distant supervision \cite{mintz2009distant} provide a way to augment the data, but these methods rely heavily on outside knowledge resources. The DS-based entity set expansion might introduce noisy instances, which potentially leads to the semantic drift problem \cite{shi2014probabilistic}. 
Ideally, we would wish to overcome these two problems, increasing the diversity of training data without external resources, and adapting our approach to any datasets. We solve the first problem by data transformation operations inside the dataset. Our method only uses the ground truth entities in the training set, which naturally reduces the influence of noisy instances. For the second problem, we use a bilateral network to enhance the learning representation, which can achieve better results on different datasets.

The context pattern provides semantics for inferring the entity slot, i.e., from ``{\it \underline{Germany} imported 47000 sheep from \underline{Britain}}" we got a context pattern ``A imported 47000 sheep from B" which implies that A and B are locations. If this sentence becomes ``{\it \underline{America} imported 47000 sheep from \underline{Britain}}", the appearance of {\it America} is also reasonable, but a person cannot appear in these placeholders. We refer to these rules as context pattern entailment, which can be emphasized and generalized by increasing entity-context diversity during model training. The data augmentation technique \cite{devries2017dataset} aims to apply a wide array of transformations to synthetically expand a training set. This paper proposes two innovative data augmentation methods on the input stage. Compared with the distant supervision, our approach does not rely on additional knowledge bases since our approach can inherently and proactively enhance low resource datasets.

We conduct experiments on five languages, including the English, German, Spanish, Dutch and Chinese, and biomedical domain. Our bilateral network achieves good performance. The main contributions of this paper can be summarized below.

(i) We augment the learning representation by increasing entity-context diversity. Our method can be applied to any datasets almost without any domain-specific modification.

(ii) To improve the versatility of our approach, we present the bilateral network to integrate the baseline and augmented representations of two sub-networks.

\section{Related Work}

The CNN based \cite{collobert2011natural}, LSTM based \cite{lample2016neural} and hybrid (i.e., LSTM-CNNs \cite{ma2016end,chiu2015named}) models resolve this task in the sequence labeling manner. Yang et. al \cite{yang2018design} build a neural sequence labeling framework\footnote{\url{https://github.com/jiesutd/NCRFpp}} to reproduce the state-of-the-art models, while we build a portable framework and also conduct experiments in different languages and domains. Yang et. al \cite{yang2017transfer} use cross-domain data and transfer learning to improve model performance.

ELMo \cite{peters2018deep} and BERT \cite{devlin2018bert} enhance the representations by pre-training language models. BERT randomly masks some words to train a masked language model, while our data augmentation is a constrained entity-context expansion. Our approach aims to retain more entity type information in the context representation. CVT \cite{clark2018semi} proposes a semi-supervised learning algorithm that uses the labeled and unlabeled data to improve the representation of the Bi-LSTM encoder.

The data augmentation method can be carried out on two stages, the raw input stage \cite{saito2017improving} and the feature space \cite{devries2017dataset}. Data augmentation paradigm has been well addressed in computer vision research, but receives less attention in NLP. Shi et. al \cite{shi2014probabilistic} propose a probabilistic Co-Bootstrapping method to better define the expansion boundary for the web-based entity set expansion. Our approach is designed to enhance the entity-context diversity of the training data without changing the entity boundary, which naturally reduces the impact of noisy instances.

The proposed framework is flexible and easy to expand. We can further consider the structural information \cite{hu2019gsn} and build the model in a paradigm of continual learning \cite{hu2018overcoming}.

\section{Methods}
\subsection{Model Overview}
\label{overview}
An NER pipeline usually contains two stages, predicting label sequence and extracting entities. Firstly, this model converts the textual input into the most likely label sequence $y^*=\arg\max\limits_{y\in Y(z)}p(y|z)$, where z and $Y(z)$ denote the textual sequence and all possible label sequences. Secondly, the post-processing module converts the label sequence into human-readable entities. The sequence labeling neural network usually contains three components for word representations, contextual representations and sequence labeling respectively.

{\bf Word representations.} This component projects each token to a $d$-dimensional vector which is composed of the word embedding and character level representation. The word embedding can be a pre-trained \cite{hu-etal-2016-different} or randomly initialized fixed-length vector.
The character level representation can be calculated by a CNN or RNN \cite{yang2018design}, and the character embeddings are randomly initialized and jointly trained.

\begin{figure*}[!h]
\centering
\includegraphics[width=3.2in]{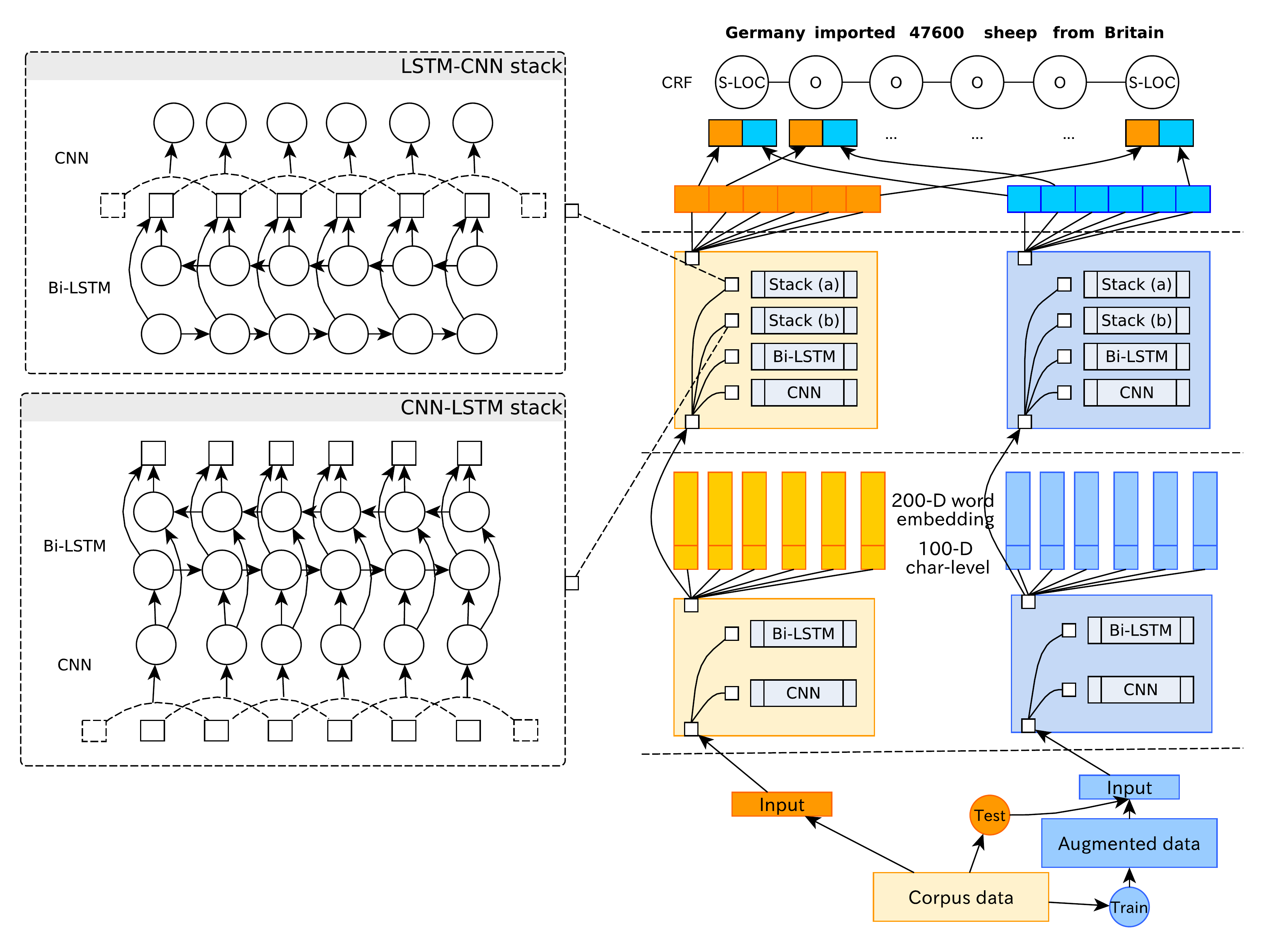}
\caption{An overview of bilateral architecture}
\label{fig_arch}
\end{figure*}

{\bf Contextual representations.} This component can generate contextual representations using CNN or RNN. Besides, our model can use different stack components to extract features, as shown in the left part of Figure \ref{fig_arch}. The major difference between these stack components is the way they extract local features. In the LSTM-CNN stack, the CNN extracts the local contextual features from the hidden state of Bi-LSTM, while in the CNN-LSTM stack the CNN extracts the local features from the word vectors and the Bi-LSTM uses the context of local features.

{\bf Sequence labeling.} This component outputs the probability of each token and selects the most likely label sequence as the final result. This paper adopts the conditional random field (CRF) \cite{lafferty2001conditional} to consider the transition probability between labels.
\begin{align}
\label{log_like}
p(y|z;W,b)&=\frac{\prod\limits_{i=1}^{n}\exp(W_{y_{i-1}y_i}^Tz_i+b_{y_{i-1}y_i})}{\sum\limits_{y'\in Y(z)}\prod\limits_{i=1}^{n}\exp(W_{y'_{i-1}y'_i}^Tz_i+b_{y'_{i-1}y'_i})} 
\end{align}
where $\{[z_i,y_i]\}, i=1,2...n$ represents the $i$-th word $z_i$ and the $i$-th label $y_i$ in the input sequence respectively. $Y(z)$ denotes all the possible label sequences for the input sequence $z$. $W$ and $b$ are weight matrix and bias vector, in which $W_{y_{i-1},y_i}$ and $b_{y_{i-1},y_i}$ are the weight vector and bias corresponding to the successive labels $(y_{i-1},y_i)$. $p(y|z;W,b)$ is the probability of generating this tag sequence over all possible tag sequences.

During the training process, the model parameters are updated to maximize the log-likelihood $ L(W,b)$. For prediction, the decoder will find the optimal label sequence that can maximize the log-likelihood $L(W,b)$ through the Viterbi algorithm. 
\begin{align}
L(W,b)&=\sum\limits_{j}\log p(y^{j}|z^{j};W,b) \\
\label{decode_crf}
y^*&=\arg\max\limits_{y\in Y(z)}p(y|z;W,b)
\end{align}
\subsection{Data Augmentation}
\label{augment}

\begin{figure}[h]
\centering
\subfloat[Sentence-centric augmentation]{
\includegraphics[width=2.5in]{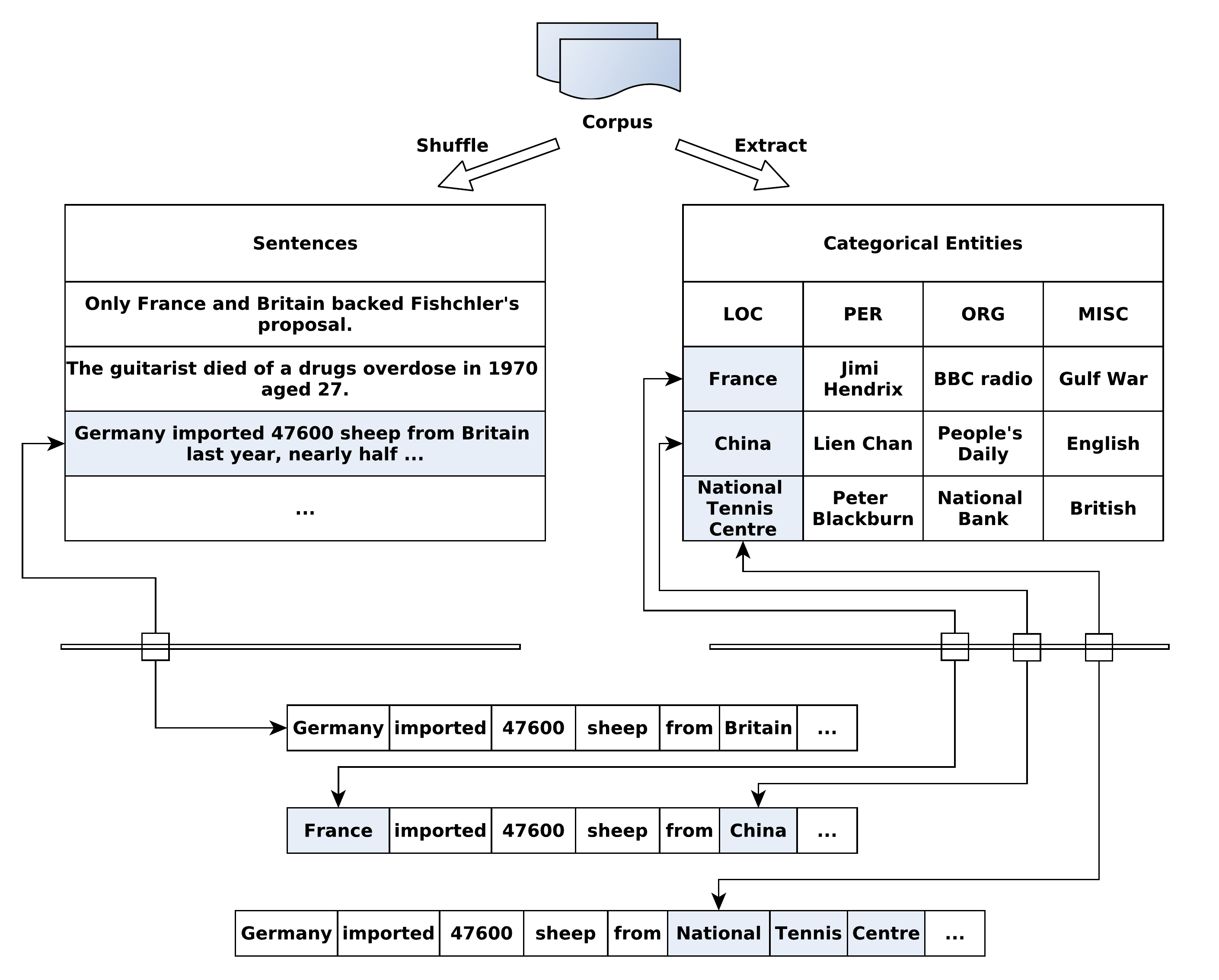}
}
\subfloat[Entity-centric augmentation]{
\includegraphics[width=2.5in]{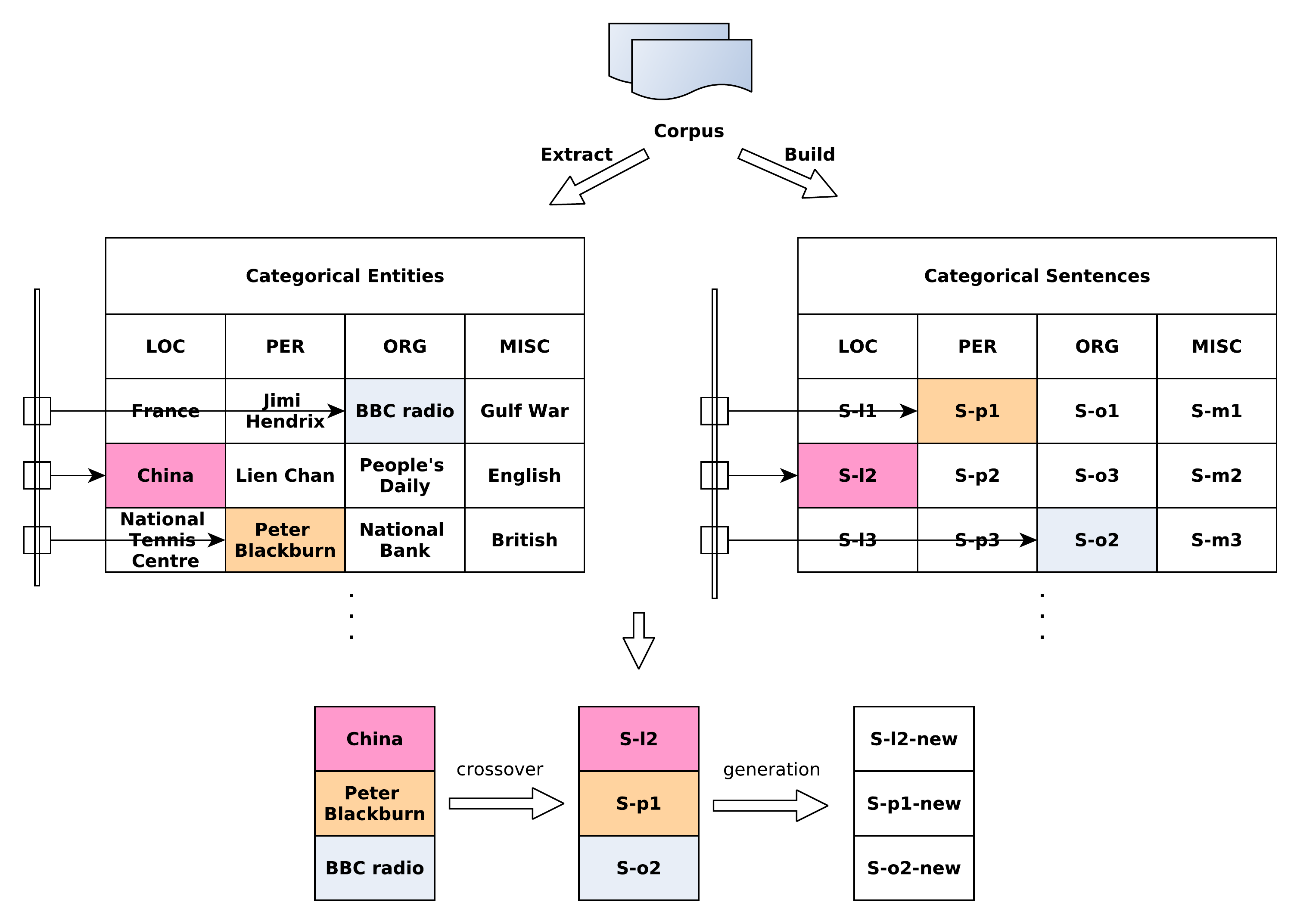}
}
\label{augments}
\caption{The schematic diagram of data augmentation operations where the black arrows represent the random selectors}
\label{shishi}
\end{figure}

{\bf Sentence-centric augmentation (SCA).} As shown in Figure \ref{shishi}(a), this method enhances the context representation by increasing entity diversity. This operation augments the entity distribution of any sample. We generate augmented sentences as follows:

1. Extract the categorical entity glossary $E=\{E_1, E_2, ..., E_c\}$ based on the original corpus $S$, where $c$ is class number. An entity may be composed of multiple words, so we needs to convert label sequence into complete entities.

2. Resample the sentence $s_i\sim\mbox{Uniform}(S)$ and light up each entity slot $a_{(i,j)} \sim \mbox{Bernoulli}(p)$. $p$ (0.5 to 0.9) is chosen according to different datasets.

3. Replace the lighted entities $a_{(i,k)} \in E_{j}$ with $\hat{a}_{(i,k)} \sim \mbox{Binomial}(E_{j}\backslash\{a_{(i,k)}\})$. This is a crossover operation.

{\bf Entity-centric augmentation  (ECA).} In SCA, we can control the augmentation for context, but the entity control is not easy. As shown in Figure \ref{shishi}(b), ECA enhances the entity representation by increasing context diversity. This operation augments the sentence distribution of an entity. We can better control the augmentation for entities.

1. Extract the categorical entity glossary $E=\{E_1, E_2, ..., E_c\}$ from the training data.

2. Build the categorical sentence set $S=\{S_1, S_2, ..., S_c\}$. The main idea is to classify the training samples according to the type mention of entities. Let $e_{(i,1)} \in E_i$ and $e_{i,1} \in s_{j}$, then $s_j \in S_i$. In Figure \ref{shishi}(b), the S-li denotes the $i$-th sentence containing at least one l ({\bf L}OC) entity.

3. Sample an entity $e_{(i,1)}$ with a probability of $p=F(e_{(i,1)})/F(E_i)$ where $F(\cdot)$ is the frequency. Sample a sentence $s_i \sim \mbox{Uniform}(S_i)$, and then perform the crossover operation.

If we iteratively substitute entities once a time in SCA, it behaves like ECA to some extent, but the distribution of entity and context is different. The basis of the two augmentation methods is different, because ECA can be extended to augment the vertex representation of a knowledge graph, while SCA focuses on text-level expansion. Here we use a random selector to simplify the augmentation process and make this work widely available. But this may lead to some noisy samples, e.g., ``Germany imported 47600 sheep from national tennis centre''. Quality control is critical when faced with data in a specific domain and is reserved for future work. Our approach actively extracts entities from within the dataset, thus not relying on the external resources. Our approach can be generally applied to augment the entity-context diversity for any datasets. 

\subsection{Bilateral architecture}
As shown in Figure \ref{fig_arch}, the bilateral architecture is composed of a baseline network on the left side and an augmented network on the right side. In the word and contextual representation layers, each sub-network can optionally use the Bi-LSTM, CNN and mixed stack layers to form flexible network combinations. This strategy can generate at least 64 ($(2\times4)^n, n\geq 2$, where $n$ is the number of sub-networks) types of bilateral networks. We compare different layer stack networks and their combinations in experiments. The data augmentation operations are only activated during the model training.

The inputs to the left and right sub-networks are different, so they form the function that adapts to different patterns. The right sub-network is more generalized, but the weakness is that it may generate noisy samples. This bilateral network also supports the joint training, but the inputs to both sides are the same. The bilateral network is a case of the multi-lateral network which also can be easily extended by adding more sub-networks. This paper adopts the IOBES scheme \cite{ratinov2009design}. 
The outputs of bilateral sub-networks are concatenated as the input to the final CRF layer. 
\subsection{Training Procedure}
We introduce two training methods, the separate training and the joint training. The separate training contains three steps.

(1) We train the left sub-network of Figure \ref{fig_arch} (freezing the right sub-network) using the human-annotated data. The left and the right sides represent the baseline and augmented models respectively. The outputs of the two sub-networks share the same CRF layer. In this step, the CRF layer only accepts the output of the baseline network. The output and the gradient of the augmented network are masked so the parameters of the augmented network are not updated. The baseline network learns the original features of the training data.

(2) Then, we train the right sub-network (freezing the left sub-network) with the human-annotated and the augmented data. The augmented data is generated dynamically based on the algorithm in subsection \nameref{augment}. Contrary to step (1), the CRF layer only accepts the output of the augmented network. This step also updates the weights of the full connection layer before the CRF layer. The augmented network enhances representations by increasing entity-context diversity.

(3) We retrain the last CRF layer (freezing all the components before the CRF layer) with the human-annotated data to fuse the representation. In this step, the functions of two sub-networks are kept, so the outputs of them are concatenated to form the rich representation from different perspectives.

We refer to the step (1) and (2) as the pre-training and step (3) as the fine-tuning. The separate training method can form two functional sub-networks, each of which retains its own characteristics.

For the joint training, we input the same sample into two sub-networks. Although the joint training can update the parameters simultaneously, the separate training achieved better results. This is because the separate training accepts different sentences in different sub-networks. The separate training retains the functionality of each sub-network, and features can be extracted independently from different perspectives, while the joint training processes the same task and focuses on extending layer width, so it did not fully extract diverse features.

\section{Experiments}
\subsection{Dataset and Evaluation}
{\bf Different languages.} For different languages, we adopt the CoNLL-2002 \cite{DBLP:conf/conll/Sang02} and CoNLL-2003 \cite{tjong2003introduction} datasets which are annotated with four types of entity, location (LOC), organization (ORG), person (PER), miscellaneous (MISC) in English, German, Dutch, Spanish. The Chinese dataset \cite{xu2017discourse} is a discourse-level dataset from hundreds of Chinese literature articles where seven types of entities (Thing, Person, Location, Time, Metric, Organization, Abstract) are annotated.

\noindent{\bf Biomedical field.} For the biomedical NER, we use the SCAI corpus which is provided by the Fraunhofer Institute for Algorithms and Scientific Computing. We focused on the International Union of Pure and Applied Chemistry (IUPAC) names (e.g., adenosine 3',5'-(hydrogen phosphate)) like the ChemSpot \cite{rocktaschel2012chemspot}. The second dataset is the GELLUS corpus \cite{kaewphan2015cell} which annotates cell line names in 1,212 documents drawn from the biomedical literature in the PubMed and PMC archives.

\subsection{Results on NER for different languages}
We use (C) and (W) to represent the {\bf c}haracter level input and {\bf w}ord level input respectively, i.e., the (C)CNN and (W)LSTM denote this model use the CNN to accept the input of character embeddings and the Bi-LSTM to accept the input of word embeddings respectively.

\subsubsection{English}
\begin{table}[h]
\centering
\caption{Results of different network combinations on the CoNLL-2003 English dataset}
\label{tab:report}
\resizebox{0.9\textwidth}{!}{
\begin{tabular}{|c|c|c|c|c|c|c|}
\hline
{\bf Layers}    & \multicolumn{6}{c|}{{\bf Models}}                           \\ \hline
Character input  & \multicolumn{3}{c|}{LSTM} & \multicolumn{3}{c|}{CNN}  \\ \hline
Word input  & LSTM  & CNN   & Stack (a) & LSTM  & CNN   & Stack (b) \\ \hline
Baseline  & 91.00$\pm$0.04 $^{a}$ & 89.89$\pm$0.06 & 90.99$\pm$0.06 & 90.95$\pm$0.06 $^{b}$& 90.13$\pm$0.04 & 90.15$\pm$0.04          \\ \hline
Augment   & 90.87$\pm$0.11 & 90.01$\pm$0.08 & 91.10$\pm$0.05 & 91.06$\pm$0.06 & 90.32$\pm$0.06 & 89.67$\pm$0.07     \\ \hline
Baseline+Baseline & 91.02$\pm$0.05  & 89.85$\pm$0.03 & 90.99$\pm$0.05    & 90.98$\pm$0.04 & 90.09$\pm$0.03 & 90.19$\pm$0.04 \\ \hline
Baseline+Augment & {\bf 91.36$\pm$0.08 } & 90.24$\pm$0.09 & {\bf 91.47$\pm$0.06 }  & 91.14$\pm$0.07 & 90.52$\pm$0.06 & 90.54$\pm$0.05      \\ \hline
\end{tabular}
}
\end{table}

Table \ref{tab:report} shows the results of 24 models on the English NER. This paper reproduces the baseline models and tests the stack models. To eliminate the influence of random factors we ran the experiments three times. $^{a}$ and $^{b}$ denote the models of (C)LSTM-(W)LSTM-CRF \cite{lample2016neural} and (C)CNN-(W)LSTM-CRF \cite{ma2016end} models respectively. In some cases, our baseline models slightly underperform (0.1$\sim$0.2\% F1) than the corresponding prototypes, but the bilateral models (Baseline+Augment) achieve better performances than the prototypes of \cite{lample2016neural,chiu2015named,collobert2011natural}. Concatenating two separate baseline models (Baseline+Baseline) almost did not change results, while the Baseline+Augment model produces better results. This demonstrates data augmentation is helpful to enhance the representations to generate better results. The entity-based data transformation paradigm has great potential to improve the performance of other tasks. More details could be found in the supplementary material\footnote{\url{https://github.com/liftkkkk/FLEXNER/blob/master/pic/appendix.pdf}\label{note1}}.

\subsection{Results on Biomedical NER}
In biomedical domain, one of the challenges is the limited size of training data. However, expanding biomedical datasets is more challenging because annotators need to design and understand domain-specific criteria, which complicates the process. There are many feature-based systems, but they cannot be used in different areas. Automatically expanding datasets is a promising way to enhance the use of deep learning models. Our method achieves good performance in the following two corpora. In the GELLUS corpus, the augmented and the bilateral models improve 5.11\% and 6.08\% F1 sore than our baseline model. This means that our approach will be a good choice in the biomedical field. 

\begin{table}[!htb]
\centering
\caption{Results of IUPAC Chemical terms and Cell lines on the SCAI chemicals corpus and the GELLUS corpus respectively}
\label{biomedical}
\begin{tabular}{|l|c|c|}\hline
{\bf Algorithm}             & {\bf SCAI}  & {\bf GELLUS} \\ \hline
OSCAR4 \cite{jessop2011oscar4}               & 57.3  & ---    \\
ChemSpot \cite{rocktaschel2012chemspot}             & 68.1  & ---    \\
CRF \cite{habibi2017deep}                  & ---   & 72.14  \\
LSTM-CRF \cite{habibi2017deep} & ---   & 73.51  \\ \hline
this work (Baseline)  & 69.08 & 78.78  \\ 
this work (Baseline$\times$2) & 69.06 & 78.80 \\
this work (Augment)   & {\bf 69.98} & 83.89  \\
this work (Bilateral) & 69.79 & {\bf 84.86} \\ \hline
\end{tabular}
\end{table}

Due to space constraints, extensive discussions and case studies will be introduced in the supplementary material\textsuperscript{\ref{note1}}.

\section{Conclusion}

This paper introduces a portable NER framework FlexNER which can recognize entities from textual input. We propose a data augmentation paradigm which does not need external data and is straightforward. We augment the learning representation by enhancing entity-context diversity. The layer stacks and sub-network combinations can be commonly used in different datasets to provide better representations from different perspectives.

It seems effortless to extend this framework to the multilingual NER research since we can use different sub-networks to learn different languages and then explore the interaction among them. Data quality control is an important task that seems to improve the learning process. Besides, this method is potential to be used in low-resource languages and may benefit in other entity related tasks. In the future, we also plan to apply this system to biomedical research, i.e., extracting the functional brain connectome \cite{zhu2016brain} or exploring the relations between drugs and diseases.

\section{Acknowledgement}
This study is supported by the Strategic Priority Research Program of Chinese Academy of Sciences (Grant No.XDB32070100). 

%
%
%
\bibliographystyle{splncs04}
\bibliography{mybibliography}
%





\end{document}